\documentclass{article} % For LaTeX2e
\usepackage{iclr2021_conference,times}

\usepackage{amsmath,amsfonts,bm}

\def\eqref#1{equation~\ref{#1}}
\def\1{\bm{1}}

\DeclareMathAlphabet{\mathsfit}{\encodingdefault}{\sfdefault}{m}{sl}
\SetMathAlphabet{\mathsfit}{bold}{\encodingdefault}{\sfdefault}{bx}{n}

\usepackage{hyperref}
\usepackage{url}
\usepackage{graphicx}
\usepackage{multirow}
\usepackage{booktabs}
\usepackage{enumitem}

\usepackage{xcolor}
\definecolor{donghan}{RGB}{243, 101, 2}
\definecolor{Red}{RGB}{255, 3, 13}
\definecolor{Plum}{RGB}{142, 69, 133}

\title{JAKET: Joint Pre-training of Knowledge Graph and Language Understanding}

\author{Donghan Yu$^{1}$\thanks{Equal contribution. Work done while the first author was an intern at Microsoft.} , Chenguang Zhu$^{2}$\footnotemark[1] , Yiming Yang$^{1}$, 
Michael Zeng$^2$\\
$^{1}$Carnegie Mellon University\\
\texttt{\{dyu2,yiming\}@cs.cmu.edu}\\
$^{2}$Microsoft Cognitive Services Research Group\\
\texttt{\{chezhu,nzeng\}@microsoft.com}
}

\iclrfinalcopy % Uncomment for camera-ready version, but NOT for submission.
\begin{document}

\maketitle

\begin{abstract}
Knowledge graphs (KGs) contain rich information about world knowledge, entities and relations. Thus, they can be great supplements to existing pre-trained language models. However, it remains a challenge to efficiently integrate information from KG into language modeling. And the understanding of a knowledge graph requires related context. We propose a novel joint pre-training framework, JAKET, to model both the knowledge graph and language. The knowledge module and language module provide essential information to mutually assist each other: the knowledge module produces embeddings for entities in text while the language module generates context-aware initial embeddings for entities and relations in the graph. Our design enables the pre-trained model to easily adapt to unseen knowledge graphs in new domains. Experimental results on several knowledge-aware NLP tasks show that our proposed framework achieves superior performance by effectively leveraging knowledge in language understanding.
\end{abstract}

\section{Introduction}
\label{intro}
Pre-trained language models (PLM) leverage large-scale unlabeled corpora to conduct self-supervised training. They have achieved remarkable performance in various NLP tasks, exemplified by BERT \citep{bert}, RoBERTa \citep{roberta}, XLNet \citep{xlnet}, and GPT series \citep{gpt, gpt-2, gpt-3}. It has been shown that PLMs can effectively characterize linguistic patterns from the text to generate high-quality context-aware representations \citep{liu2019linguistic}. However, these models struggle to grasp world knowledge, concepts and relations, which are very important in language understanding~\citep{poerner2019bert,talmor2019olmpics}. 

Knowledge graphs (KGs) represent entities and relations in a structural way. They can also solve the sparsity problem in text modeling. For instance, a language model may require tens of instances of the phrase ``labrador is a kind of dog" in its training corpus before it implicitly learns this fact. In comparison, a knowledge graph can use two entity nodes ``labrador", ``dog" and a relation edge ``is\_a'' between these nodes to precisely represent this fact.

Recently, some efforts have been made to integrate knowledge graphs into language model pre-training. Most approaches combine token representations in PLM with representations of aligned KG entities. The entity embeddings are either pre-computed from an external source by a separate model~\citep{zhang2019ernie,knowbert}, which may not easily align with the language representation space, or directly learned as model parameters~\citep{EAE,FAE}, which will cause an over-parameterization issue due to the large number of entities. Moreover, all the previous works share a common challenge: when the pre-trained model is fine-tuned in a new domain with a previously unseen knowledge graph, it struggles to adapt to the new entities, relations and structure.

%Current approaches to integrate knowledge graphs into pre-trained language models can be categorized into two groups. The first group employs network modules to provide embeddings for entities appearing in the training text, which are then combined with text representation \citep{zhang2019ernie}. This approach ignores the structural information in the knowledge graph which could be essential in understanding the relationship between entities. The second group utilizes a graph computation module to integrate the structural information, exemplified by TransE \citep{transe} and graph neural network \citep{gat}. The produced entity embeddings are then fed into the language module. However, many KG entities are hard to understand without context. Merely treating them as individual concepts is inconsistent with the contextual understanding in the language module. Moreover, the two approaches above share a common challenge: when the pre-trained model is fine-tuned in a new domain with a previously unseen knowledge graph, it struggles to adapt to the new entities, relations and structure. \dy{This part seems not very accurate?}

Therefore, we propose JAKET, a Joint pre-trAining framework for KnowledgE graph and Text. Our framework contains a knowledge module and a language module, which mutually assist each other by providing required information to achieve more effective semantic analysis.
The knowledge module leverages a graph attention network \citep{gat} to provide structure-aware entity embeddings for language modeling. And the language module produces contextual representations as initial embeddings for KG entities and relations given their descriptive text. Thus, in both modules, content understanding is based on related knowledge and rich context. On one hand, the joint pre-training effectively projects entities/relations and text into a shared semantic latent space. On the other hand, as the knowledge module produces representations from descriptive text, it solves the over-parameterization issue since entity embeddings are no longer part of the model's parameters. %Furthermore, this design can quickly adapt the model to downstream tasks with previously unseen knowledge graph. 

In order to solve the cyclic dependency between the two modules, we propose a novel two-step language module $\text{LM}_1$ + $\text{LM}_2$. $\text{LM}_1$ provides embeddings for both $\text{LM}_2$ and KG. The entity embeddings from KG are also fed into $\text{LM}_2$, which produces the final representation. $\text{LM}_1$ and $\text{LM}_2$ can be easily established as the first several transformer layers and the rest layers of a pre-trained language model such as BERT \citep{bert} and RoBERTa \citep{roberta}. Furthermore, we design an entity context embedding memory with periodic update which speeds up the pre-training by 15x.

The pre-training tasks are all self-supervised, including entity category classification and relation type prediction for the knowledge module, and masked token prediction and masked entity prediction for the language module.

\begin{figure}
    \centering
    \includegraphics[width=12cm]{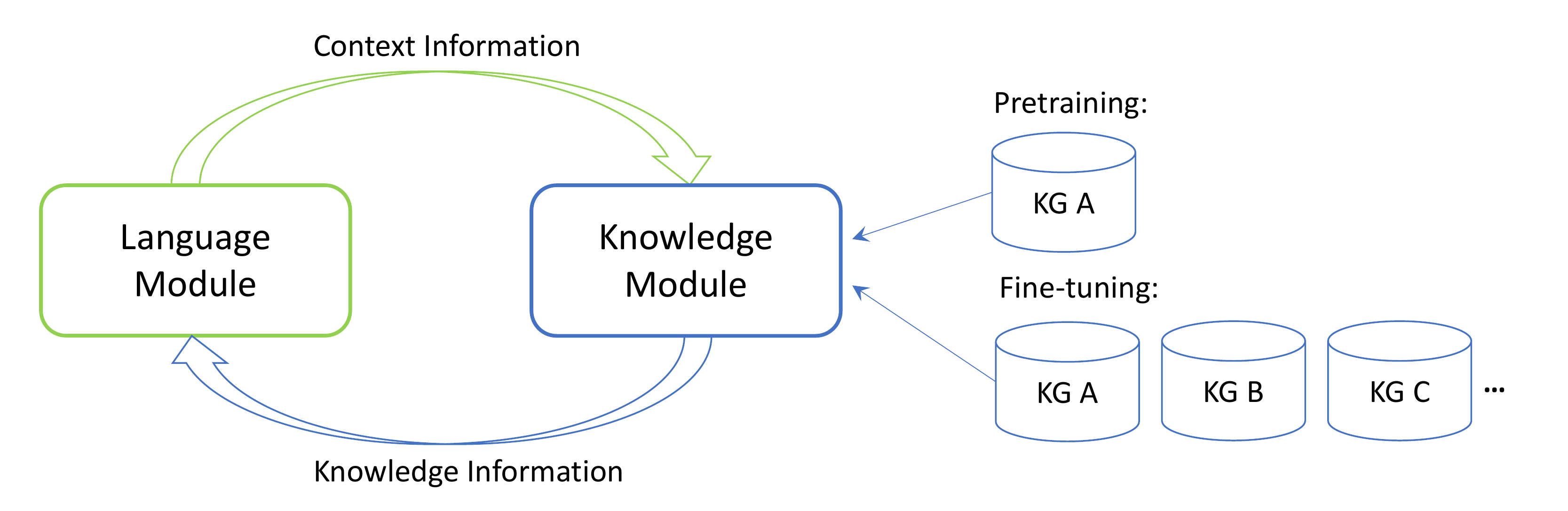}
    \caption{A simple illustration on the novelty of our proposed model JAKET.}
    \label{fig:model}
\end{figure}

A great benefit of our framework is that it can easily adapt to unseen knowledge graphs in the fine-tuning phase. As the initial embeddings of entities and relations come from their descriptive text, JAKET is not confined to any fixed KG. With the learned ability to integrate structural information during pre-training, the framework is extensible to novel knowledge graphs with previously unseen entities and relations, as illustrated in \autoref{fig:model}.

We conduct empirical studies on several knowledge-aware language understanding tasks, including few-shot relation classification, question answering and entity classification. The results show that JAKET achieves the best performance compared with strong baseline methods on all the tasks, including those with a previously unseen knowledge graph.

\section{Related Work}
\label{relate}

Pre-trained language models have been shown to be very effective in various NLP tasks, including ELMo \citep{elmo}, GPT \citep{gpt}, BERT~\citep{bert}, RoBERTa \citep{roberta} and XLNet~\citep{xlnet}. Built upon large-scale corpora, these pretrained models learn effective representations for various semantic structures and linguistic relationships. They are trained on self-supervised tasks like masked language modeling and next sentence prediction. 

Recently, a lot of efforts have been made on investigating how to integrate knowledge into PLMs~\citep{levine2019sensebert,MTB,k-bert,guu2020realm}. These approaches can be grouped into two categories:

1. Explicitly injecting entity representation into language model, where the representations are either pre-computed from external sources~\citep{zhang2019ernie,knowbert} or directly learned as model parameters~\citep{EAE,FAE}. For example, ERNIE (THU) \citep{zhang2019ernie} pre-trains the entity embeddings on a knowledge graph using TransE~\citep{transe},  while EAE~\citep{EAE} learns the representation  from pre-training objectives with all the model parameters.

2. Implicitly modeling knowledge information, including entity-level masked language modeling~\citep{sun2019ernie,shen2020exploiting}, entity-based replacement prediction~\citep{xiong2019pretrained} and knowledge embedding loss as regularization~\citep{wang2019kepler}. For example, besides token-level masked language modeling, ERNIE (Baidu) \citep{sun2019ernie} uses phrase-level and entity-level masking to predict all the masked slots. KEPLER \citep{wang2019kepler} calculates entity embeddings using a pre-trained language model based on the description text, which is similar to our work. However, they use the entity embeddings for the knowledge graph completion task instead of injecting them into language model.

Some works~\citep{ding2019cognitive,gnn-plm-cqa} investigated the combination of GNN and PLM. For example, \cite{gnn-plm-cqa} uses XLNet to generate initial node representation based on node context and feeds them into a GNN. However, these approaches do not integrate knowledge into language modeling, and they are designed for specific NLP tasks such as reading comprehension or commonsense reasoning. In comparison, we jointly pre-train both the knowledge graph representation and language modeling and target for general knowledge-aware NLU tasks.

%Moreover, None of these works explicitly model the relation information between entities except FAE~\citep{FAE}, which builds a large memory to encode triplets from KG. But unlike our proposed model, it can not capture multi-hop structure information.

% \textbf{Graph Neural Network.} Graph Neural Network (GNN) has received increasing attention in recent machine learning research. \citet{gcn} formulates a uniform Message Passing framework for most followup works. Thereafter, many extension works try to improve its effectiveness by reweighting edge weights \citep{gat}, making residual links \citep{li2019deepgcns} and incorporating multi-relation edge types \citep{rgcn,compgcn}. Some other works~\citep{ding2019cognitive,gnn-plm-cqa} investigated its combination with PLMs, but are only designed for specific NLP tasks like Reading Comprehension of Commonsense Reasoning \cz{some more details here}. Moreover, some pre-training techniques~\citep{strategies-gnn,gpt-gnn} have been proposed for GNN to improve its performance of various downstream tasks, while our work jointly pretrain GNN and language model, which is more broadly applicable.

\section{Method}
\label{method}

\begin{figure}
    \centering
    \includegraphics[width=14cm]{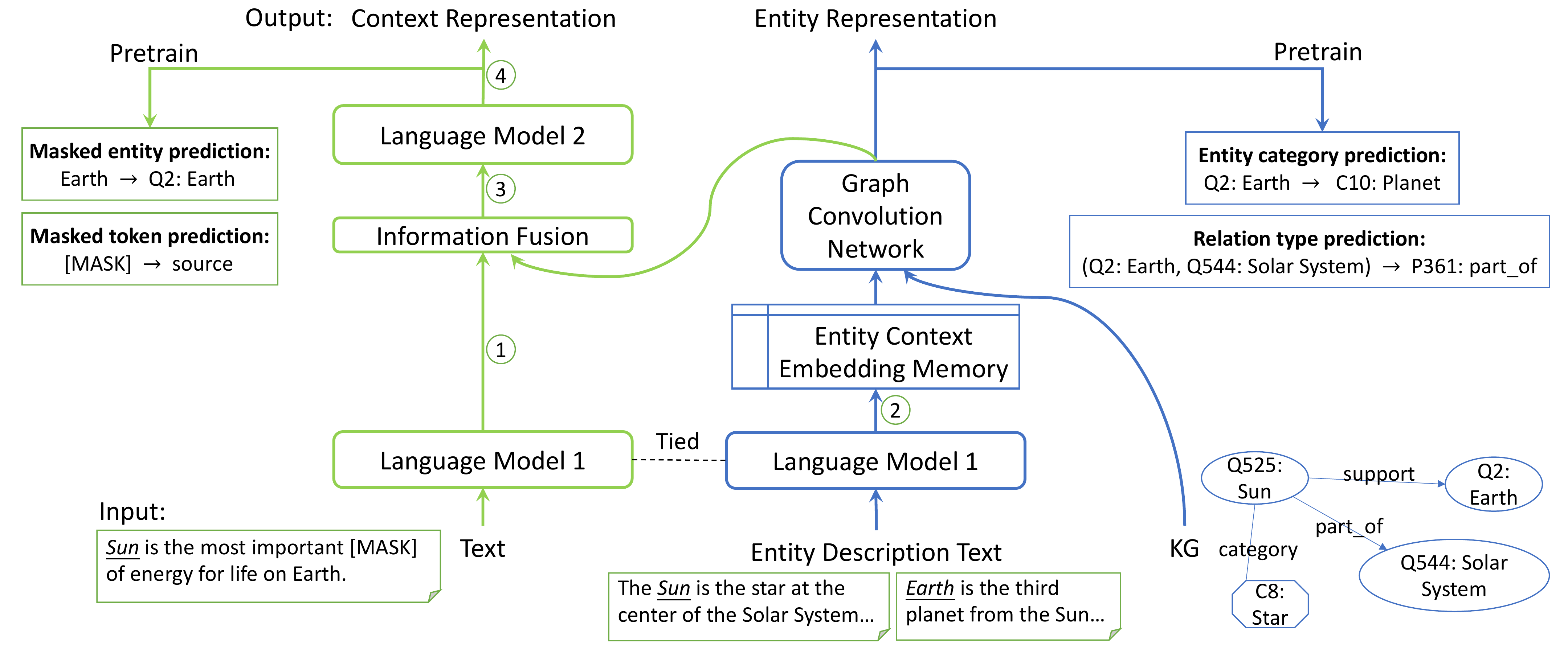}
    \caption{A demonstration for the structure of JAKET, where the language module is on the left side marked green while the knowledge module is on the right side marked blue. Symbol \textcircled{\texttt{X}} indicates the steps to compute context representations introduced in Section \ref{sec:cyclic}. ``Q\texttt{X}", ``P\texttt{X}" and ``C\texttt{X}" are the indices for entities, relations and categories in KG respectively. Entity mentions in text are underlined and italicized such as \underline{\textit{Sun}}.} %Note that the ``Earth" mention is not marked in the input text since it's masked and set as the pre-training objective.} 
    \label{fig:model2}
\end{figure}

In this section, we introduce the JAKET framework of joint pre-training knowledge graph and language understanding. We begin by defining the mathematical notations, and then present our model architecture with the knowledge module and language module. Finally, we introduce how to pre-train our model and fine-tune it for downstream tasks. The framework is illustrated in \autoref{fig:model2}.

\subsection{Definition}
A knowledge graph is denoted by $\mathcal{KG} = ( \mathcal{E}, \mathcal{R}, \mathcal{T})$, where $\mathcal{E} = \{e_1 \ldots e_N \}$ is the set of entities and $\mathcal{R} = \{r_1 \ldots r_P \}$ is the set of relations. $\mathcal{T} = \{ (e_{t_i^1}, r_{t_i^2}, e_{t_i^3}) | 1\leq i \leq T, e_{t_i^1}, e_{t_i^3}\in \mathcal{E}, r_{t_i^2} \in \mathcal{R} \}$ stands for the set of head-relation-tail triplets. $N_v=\{(r, u)|(v,r,u) \in \mathcal{T}\}$ represents the set of neighboring relations and entities of an entity $v$.

We define $\mathcal{V}=\left\{[\text{MASK}], [\text{CLS}], [\text{EOS}], w_{1} \ldots w_{V}\right\}$ as a vocabulary of tokens and the contextual text $\mathbf{x}=\left[x_{1}, x_2, \ldots, x_L\right]$ as a sequence of tokens where $x_i \in \mathcal{V}$. In the vocabulary, $[\text{MASK}]$ is the special token for masked language modeling~\citep{bert} and $[\text{CLS}], [\text{EOS}]$ are the special tokens indicating the beginning and end of the sequence. We define $F$ as the dimension of token embeddings, which is equal to the dimension of entity/relation embeddings from the knowledge graph.

The text $\mathbf{x}$ has a list of entity mentions $\mathbf{m} = [m_1, \ldots, m_M]$, where each mention $m_i=(e_{m_i}, s_{m_i}, o_{m_i})$: $e_{m_i}$ is the corresponding entity and $s_{m_i}, o_{m_i}$ are the start and end index of this mention in the context. In other words, $[x_{s_{m_i}}, \ldots, x_{o_{m_i}}]$ is linked with entity $e_{m_i}$\footnote{We do not consider discontinous entity mentions in this work.}. We assume the span of mentions are disjoint for a given text sequence.

As entities in the knowledge graph are represented by nodes without context, we use \textit{entity description text} to describe the concept and meaning of entities. For each entity $e_i$, its description text $\mathbf{x}^{e_i}$ describes this entity. The mention of $e_i$ in $\mathbf{x}^{e_i}$ is denoted as $m^{e_i}=(e_i, s_i^e, o_i^e)$, similarly defined as above. For instance, the description text for the entity ``sun'' can be ``[CLS] The Sun is the star at the center of the Solar System [EOS]''. Then the mention is $m^{Sun}=(Sun, 3, 3)$. If there are multiple mentions of $e_i$ in its description text, we choose the first one.
If there's no mention of $e_i$ in its description text, we set $s_i^e=o_i^e=1$. Similarly, we define \textit{relation description text} as the text that can describe each relation.% and the mention of relation $r_i$ is denoted as $m^{r_i}=(r_i, s_i^r, o_i^r)$.

%where $s_i^e, o_i^e$ are the starting index and end index of the corresponding mention of entity $e_i$.

%Each token in context and each entity is represented by a vector. We denote $E$ as the entity representation matrix and $Z$ as the token representation matrix.

% $\mathcal{D} = \{ \mathbf{d}_1 \ldots \mathbf{d}_N \}$ where $\mathbf{d}_i$ is the specific context describing entity $e_i$. In our work, each entity has at most one description context.

% \begin{figure}
%     \centering
%     \includegraphics[width=14cm]{fig/model2.pdf}
%     \caption{Caption}
%     \label{fig:my_label}
% \end{figure}

\subsection{Knowledge Module}

The goal of the knowledge module ($\text{KM}$) is to model the knowledge graph to generate knowledge-based entity representations.

To compute entity node embeddings, we employ the graph attention network (GAT) \citep{gat}, which uses the self-attention mechanism to specify different weights for different neighboring nodes. However, the vanilla GAT is designed for homogeneous graphs with single-relation edges. To leverage the multi-relational information, we adopt the idea of composition operator \citep{compgcn} to compose entity embeddings and relation embeddings. In detail, in the $l$-th layer of LM, we update the embedding $E_{v}^{(l)}$ of entity $v$ as follows:
\begin{align}
\label{eq:updateE}
E^{(l)}_{v} & =\text{LayerNorm} \left( \bigoplus_{k=1}^{K} \sigma \left(\sum_{(r,u) \in \mathcal{N}_{v}} \alpha_{v,r,u}^{k} W^{k} f(E^{(l-1)}_{u}, R_r) \right) + E^{(l-1)}_{v} \right) \\
\alpha_{v,r,u}^{k} &  = \frac{ \exp \left(\text { LeakyReLU }\left(\mathbf{a}^{T}\left[W^k E^{(l-1)}_{v} \oplus W^k f(E^{(l-1)}_{u}, R_r)\right]\right)\right)}{\sum_{(r^{\prime},u^{\prime}) \in \mathcal{N}_{v}} \exp \left(\text { LeakyReLU }\left(\mathbf{a}^{T}\left[W^k E^{(l-1)}_{u} \oplus W^k f(E^{(l-1)}_{u^{\prime}}, R_{r^{\prime}})\right]\right)\right)}
\end{align}
where $\bigoplus$ means concatenation and $K$ is the number of attention heads. $W^k$ is the model parameter and $R_r$ is the embedding of relation $r$. Note that the relation embeddings are shared across different layers. The function $f(\cdot, \cdot): \mathbb{R}^F \times \mathbb{R}^F \rightarrow \mathbb{R}^F$ merges a pair of entity and relation embeddings into one representation. Here, we set $f(x, y)=x+y$ inspired by TransE~\citep{transe}. More complicated functions like MLP network can also be applied. 

The initial entity embeddings $E^{(0)}$ and relation embeddings $R$ are generated from our language module, which will be introduced in Section \ref{sec:languagemodule}. Then, the output entity embeddings from the last GAT layer are used as the final entity representations $E^{\text{KM}}$. Note that the knowledge graph can be very large, making the embedding update over all the entities in Equation (\ref{eq:updateE}) not tractable. 
Thus we follow the minibatch setting \citep{graphsage}: given a set of input entities, we perform neighborhood sampling to generate their multi-hop neighbor sets and we compute representations only on the entities and relations that are necessary for the embedding update.

% We define the whole process as:
% \begin{align}
% E^{kg} = \text{GNN}(E^{context}, \mathcal{KG})
% \label{eq:KMod}
% \end{align}

% \begin{align}
% \mathcal{L}_{c} = -\sum_{i\in \mathcal{Y}_{L}} \sum_{j=1}^C Y_{ij}ln \widehat{Y}_{ij} 
% \end{align}
% where $\mathcal{L}_{c}$ is the loss of entity category prediction. $\mathcal{Y}_{L}$ is the labeled entity set. $C$ is the number of categories. $Y_{ij} = 0$ means the true label of entity $i$ is $j$ while $\widehat{Y}_{ij}$ is the model predicted value.

% \begin{align}
% \mathcal{L}_{r} = -\sum_{i\in \mathcal{Y}_{L}} \sum_{j=1}^C Y_{ij}ln \widehat{Y}_{ij}
% \end{align}
% The total loss of knowledge module pretraining is $\mathcal{L}_a + \mathcal{L}_r$.

\subsection{Language Module}
\label{sec:languagemodule}

The goal of the language module (LM) is to model text data and learn context-aware representations. The language module can be any model for language understanding, e.g. BERT \citep{bert}. In this work, we use pre-trained model RoBERTa-base \citep{roberta} as the language module.

\subsection{Solving the cyclic dependency}
\label{sec:cyclic}

In our framework, the knowledge and language modules mutually benefit each other: the language module $\text{LM}$ outputs context-aware embedding to initialize the embeddings of entities and relations in the knowledge graph given the description text; the knowledge module (KM) outputs knowledge-based entity embeddings for the language module.

However, there exists a cyclic dependency which prevents computation and optimization in this design. To solve this problem, we propose a decomposed language module which includes two language models: $\text{LM}_1$ and $\text{LM}_2$. We employ the first 6 layers of RoBERTa as $\text{LM}_1$ and the remaining 6 layers as $\text{LM}_2$. The computation proceeds as follows:
\begin{enumerate}
    \item $\text{LM}_1$ operates on the input text $\mathbf{x}$ and generates contextual embeddings $Z$.
    \item $\text{LM}_1$ generates initial entity and relation embeddings for $\text{KM}$ given description text.
    \item KM produces its output entity embeddings to be combined with $Z$ and sent into $\text{LM}_2$.
    \item $\text{LM}_2$ produces the final embeddings of $\mathbf{x}$, which includes both contextual and knowledge information.
\end{enumerate}

In detail, in step 1, suppose the context $\mathbf{x}$ is embedded as $X^{embed}$. $\text{LM}_1$ takes $X^{embed}$ as input and outputs hidden representations $Z = \text{LM}_1(X^{embed})$.

In step 2, suppose $\mathbf{x}^{e_j}$ is the \textit{entity description text} for entity $e_j$, and the corresponding mention is $m^{e_j} = (e_j, s_j^e, o_j^e)$. $\text{LM}_1$ takes the embedding of $\mathbf{x}^{e_j}$ and produces the contextual embedding $Z^{e_j}$. Then, the average of embeddings at position $s_j^e$ and $o_j^e$ is used as the initial entity embedding of $e_j$, i.e. $E^{(0)}_j  = (Z_{s_j^e}^{e_j} + Z_{o_j^e}^{e_j})/2$.
%Since the entity embeddings are generated from contextual embeddings, we call them \textit{entity context embeddings} and denote them as $E^{context}$ in the later part of our paper. 
The knowledge graph relation embeddings $R$ are generated in a similar way using its description text. 

In step 3, $\text{KM}$ computes the final entity embeddings $E^{\text{KM}}$, which is then combined with the output $Z$ from $\text{LM}_1$. In detail, suppose the mentions in $\mathbf{x}$ are $\mathbf{m}=[m_1, \ldots, m_M]$. $Z$ and $E^{\text{KM}}$ are combined at positions of mentions:
\begin{align}
Z_k^{merge}=\left\{
\begin{array}{ccl}
Z_k + E_{e_{m_i}}^{\text{KM}}       &      & \text{if }  \exists i \text{ s.t. } s_{m_i} \leq k  \leq o_{m_i}  \\
Z_k    &      & \text{otherwise}
\end{array} \right.
\end{align}
where $E_{e_{m_i}}^{\text{KM}}$ is the output embedding of entity $e_{m_i}$ from $\text{KM}$.

We apply layer normalization~\citep{layernorm} on $Z^{merge}$: $Z^{\prime} = \text{LayerNorm}(Z^{merge})$. Finally, $Z^{\prime}$ is fed into $\text{LM}_2$.

In step 4, $\text{LM}_2$ operates on the input $Z^{\prime}$ and obtains the final embeddings $Z^{\text{LM}} = \text{LM}_2(Z^{\prime})$. The four steps are marked by symbol \textcircled{\texttt{X}} in \autoref{fig:model2} for better illustration.

% \cz{Draw a good plot for the whole design, using the same letters as in the text. Give an example text and entity graph.}

% \cz{Move this paragraph in related work}Note that although the archtecture is similar to EAE~\citep{EAE}, our model has two key differences: 1. the first language model will provide context embeddings for entities; 2. the entity embeddings output from the knowledge module contain multi-hop graph structure information captured by GNN.

% we use the token embeddings of both the start and end indicies of the mention for entity id prediction. Then, the model predicts the entity whose embedding in E is the closest to the pseudo entity embedding. To prevent this task from being trivial, we do not update the token embeddings with entity context embeddings. Instead, we directly set $Z_i^{\prime} = Z_i$. 

\subsection{Entity Context Embedding Memory}
Many knowledge graphs contain a large number of entities. Thus, even for one sentence, the number of entities plus their multi-hop neighbors can grow exponentially with the number of layers in the graph neural network. As a result, it's very time-consuming for the language module to compute context embeddings based on the description text of all involved entities in a batch on the fly.%\footnote{In our experiment, this step would take about 95$\%$ of the time in pretraining.}. 

To solve this problem, we construct an entity context embedding memory, $E^{context}$, to store the initial embeddings of all KG entities. Firstly, the language module pre-computes the context embeddings for all entities and place them into the memory. The knowledge module only needs to retrieve required embeddings from the memory instead of computing them, i.e. $E^{(0)} \leftarrow E^{context}$.

However, as embeddings in the memory are computed from the ``old'' (initial) language module while the token embeddings during training are computed from the updated language module, there will be an undesired discrepancy. Thus, we propose to update the whole embedding memory $E^{context}$ with the current language module every $T(i)$ steps, where $i$ is the number of times that the memory has been updated (starting from 0). $T(i)$ is set as follows:
\begin{align}
T(i) = \min(I_{init} * a^{\lfloor i/r \rfloor}, I_{max})
\end{align}
where $I_{init}$ is the initial number of steps before the first update and $a$ is the increasing ratio of updating interval. $r$ is the number of repeated times of the current updating interval. $I_{max}$ is the maximum number of steps between updates. In our experiments, we set $I_{init}=10, a=2, r=3, I_{max}=500$, and the corresponding squence of $T$ is $[10,10,10,20,20,20,40,40,40,\ldots,500,500]$. Note that we choose $a > 1$ because the model parameters usually change less as training proceeds.

Moreover, we propose a momentum update to make $E^{context}$ evolve more smoothly. Suppose the newly calculated embedding memory by LM is $E^{context}_{new}$, then the updating rule is:
\begin{align}
E^{context} \leftarrow mE^{context} + (1-m)E_{new}^{context},
\end{align}
where $m\in [0,1)$ is a momentum coefficient which is set as $0.8$ in experiment.  

This memory design speeds up our model by about 15x during pre-training while keeping the effectiveness of entity context embeddings. For consideration of efficiency, we use relation embeddings only during fine-tuning.

\subsection{Pre-training}

During pre-training, both the knowledge module and language module are optimized based on several self-supervised learning tasks listed below. The examples of all the training tasks are shown in \autoref{fig:model2}.

At each pre-training step, we first sample a batch of root entities and perform random-walk sampling %\footnote{To perform random walk sampling, we transform the knowledge graph into undirected unweighted homogeneous graph by discarding the relation information.}
on each root entity. The sampled entities are fed into KM for the following two tasks. 

\textbf{Entity category prediction.} The knowledge module is trained to predict the category label of entities based on the output entity embeddings $E^{\text{KM}}$. The loss function is cross-entropy for multi-class classification, denoted as $\mathcal{L}_c$. 

\textbf{Relation type prediction.} KM is also trained to predict the relation type  between a given entity pair based on $E^{\text{KM}}$. The loss function is cross-entropy for multi-class classification, denoted as $\mathcal{L}_r$. 

Then, we uniformly sample a batch of text sequences and their entities for the following two tasks.

\textbf{Masked token prediction.} Similar to BERT, We randomly mask tokens in the sequence and predict the original tokens based on the output $Z^{\text{LM}}$ of language module. We denote the loss as $\mathcal{L}_t$.

\textbf{Masked entity prediction.} The language module is also trained to predict the corresponding entity of a given mention. For the input text, we randomly remove $15\%$ of the mentions $\mathbf{m}$. Then for each removed mention $m_r = (e_r, s_r, o_r)$, the model predicts the masked entity $e_r$ based on the mention's embedding. In detail, it predicts the entity whose embedding in $E^{context}$ is closest to $q=g((Z^{\text{LM}}_{s_r} + Z^{\text{LM}}_{o_r})/2)$, where $g(x)=\text{ReLU}(xW_1)W_2$ is a transformation function.
Since the number of entities can be very large, we use $e_r$'s neighbours and other randomly sampled entities as negative samples. The loss function $\mathcal{L}_{e}$ is cross entropy based on the inner product between $q$ and each candidate entity's embedding. \autoref{fig:model2} shows an concrete example, where the mention ``Earth" is not marked in the input text since it's masked and the task is to link the mention ``Earth" to entity ``Q2: Earth".

\subsection{Fine-tuning}
During fine-tuning, our model supports using either the knowledge graph employed during pre-training or a novel custom knowledge graph with previously unseen entities\footnote{We assume the custom domain comes with NER and entity linking tools which can annotate entity mentions in text. The training of these systems is beyond the scope of this work.}. If a custom KG is used, the entity context embedding memory is recomputed by the pre-trained language module using the new entity description text. In this work, we do not update the entity context memory during fine-tuning for consideration of efficiency. We also compute the relation context embedding memory using the pre-trained language model.

% Both the context representations output by the language module and the entity representations output by the knowledge module can be used for downstream tasks, which will be shown in the experiment section. 

\section{Experiment}
\label{exp}

%We begin by introducing experimental settings including the data for pre-training, implementation details and compared baselines. Then we show the results on one downstream task  using the pre-trained knowledge graph and two tasks using novel knowledge graphs with unseen entities.

\subsection{Basic Settings}

\textbf{Data for Pre-training.}
We use the English Wikipedia %\footnote{https://en.wikipedia.org}
as the text corpus, Wikidata~\citep{wikidata} as the knowledge graph, and SLING~\citep{sling} to identify entity mentions. For each entity, we use the first 64 consecutive tokens of its Wikipedia page as its description text and we filter out entities without a corresponding Wikipedia page. We also remove entities that have fewer than 5 neighbors in the Wikidata KG and fewer than 5 mentions in the Wikipedia corpus. The final knowledge graph contains 3,657,658 entities, 799 relations and 20,113,978 triplets. We use the \textit{instance of} relation to find the category of each entity. In total, 3,039,909 entities have category labels of 19,901 types. The text corpus contains about 4 billion tokens. 

\textbf{Implementation Details.}
We initialize the language module with the pre-trained RoBERTa-base~\citep{roberta} model. The knowledge module is initialized randomly. Our implementation is based on the HuggingFace framework \citep{huggingface} and DGL~\citep{dgl}. 
%We use the embedding layer to obtain initial token embeddings. We employ the first 6 layers of RoBERTa as $\text{LM}_1$ and the remaining 6 layers as $\text{LM}_2$. 
For the knowledge module, we use a 2-layer graph neural network, which aggregates 2-hop neighbors. The number of sampled neighbors in each hop is 10. More details are presented in the Appendix.

\textbf{Baselines.} We compare our proposed model JAKET with the pre-trained RoBERTa-base~\citep{roberta} and two variants of our model: RoBERTa+GNN and RoBERTa+GNN+M. The two models have the same model structure as JAKET, but they are not pre-trained on our data. The entity and relation context embedding memories of RoBERTa+GNN are randomly generated while the memories of RoBERTa+GNN+M are computed by the RoBERTa.

\subsection{Downstream tasks}
\textbf{Few-shot Relation Classification}. Relation classification requires the model to predict the relation between two entities in text. Few-shot relation classification takes the $N$-way
$K$-shot setting. Relations in the test set are not seen in the training set. For each query instance, $N$ relations with $K$ supporting examples for each relation are given. The model is required to classify the instance into one of the $N$ relations based on the $N \times K$ samples. In this paper we evaluate our model on FewRel~\citep{fewrel}, which is a widely used benchmark dataset for few-shot relation classification, containing 100 relations and 70,000 instances. 

We use the pre-trained knowledge graph for FewRel as it comes with entity mentions from Wikidata knowledge graph.
To predict the relation label, we build a sequence classification layer on top of the output of LM. More specifically, we use the PAIR framework proposed by~\cite{fewrel2}, which pairs each query instance with all the supporting instances, concatenate each pair as one sequence, and send the concatenated sequence to our sequence classification model to get the score of the two instances expressing the same relation. We do not use relation embeddings in this task to avoid information leakage.

As shown in Table~\ref{tab:fewrel}, our model achieves the best results in all three few-shot settings. Comparing the results between RoBERTa and RoBERTa+GNN, we see that adding GNN with randomly generated entity features does not improve the performace. The difference between RoBERTa+GNN+M and RoBERTa+GNN demonstrates the importance of generating context embedding memory by the language module, while JAKET can further improve the performance by pre-training.

\begin{table}[]
\centering
\begin{tabular}{lccc}
\hline
      Model             & 5-way 1-shot                  & 5-way 5-shot                      & 10-way 1-shot                     \\ \hline
PAIR (BERT)$^{\star}$        & 85.7                                        & 89.5                                          & 76.8                                         \\ %\hline
PAIR (RoBERTa)     & 86.4                                          & 90.3                                        & 77.3                                          \\ %\hline
PAIR (RoBERTa+GNN)     & 86.3                                          & -                                     & -                                          \\ %\hline
PAIR (RoBERTa+GNN+M)     & 86.9                                          & -                                        & -                                          \\ %\hline
PAIR (JAKET)        &  \textbf{87.4}  &  \textbf{92.1}  &  \textbf{78.9}  \\ \hline
\end{tabular}
\caption{\label{tab:fewrel} Accuracy results on the dev set of FewRel 1.0. $\star$ indicates the results are taken from~\cite{fewrel2}. PAIR is the framework proposed by~\cite{fewrel2}.}
\end{table}

% \begin{table}[]
% \centering
% \begin{tabular}{|l|c|c|c|c|c|c|}
% \hline
%                   & \multicolumn{2}{c|}{5 way 1 shot}                      & \multicolumn{2}{c|}{5 way 5 shot}                      & \multicolumn{2}{c|}{10 way 1 shot}                     \\ \cline{2-7} 
% \multirow{-2}{*}{FewRel 1.0} & dev                         & test                     & dev                         & test                     & dev                         & test                     \\ \hline
% MTB ($\text{BERT}_{large}$)         & 90.1                        & 93.9                     &                             & 97.1                     & 83.4                        & 89.2                     \\ \hline
% PAIR (BERT)        & 85.7                        & 88.32                    & 89.5                        & 93.22                    & 76.8                        & 80.63                    \\ \hline
% PAIR (RoBERTa)     & 86.4                        & 89.32                    & 90.3                        & 93.7                     & 77.3                        & 82.49                    \\ \hline
% PAIR (KEPLER)      & -                           & 90.31                    & -                           & 94.28                    & -                           & 85.48                    \\ \hline
% PAIR (Ours)        &  87.4 & - &  92.1 &  - &  78.9 & - \\ \hline
% \end{tabular}
% \caption{Results on FewRel 1.0.}
% \end{table}

\textbf{KGQA}. The Question Answering over KG (KGQA) task is to answer natural language questions related to a knowledge graph. The answer to each question is an entity in the KG. This task requires an understanding over the question and reasoning over multiple entities and relations.

We use the vanilla version of the MetaQA~\citep{metaqa} dataset, which contains questions requiring multi-hop reasoning over a novel movie-domain knowledge graph. The KG contains 135k triplets, 43k entities and 9 relations. Each question is provided with one entity mention and the question is named as a $k$-hop question if the answer entity is a $k$-hop neighbor of the question entity. We define all the $k$-hop neighbor entities of the question entity as the candidate entities for the question. We also consider a more realistic setting where we simulate an incomplete KG by randomly dropping a triplet with a probability $50\%$. This setting is called \textit{KG-50\%}, compared with the full KG setting \textit{KG-Full}. 

For each entity, we randomly sample one question containing it as the entity's description context. We manually write the description for each relation since the number of relations is very small. We use the output embedding of [CLS] token from LM as the question embedding, and then find the entity with the closest context embedding.

As shown in Table~\ref{tab:kgqa},  RoBERTa+GNN+M outperforms RoBERTa, demonstrating the effectiveness of KM+LM structure. JAKET further improves the accuracy by 0.6\% to 2.5\% under both KG settings, showing the benefits of our proposed joint pre-training.\footnote{For fair comparison, we do not include models which incorporate a dedicated graph retrieval module~\citep{GraftNet,pullnet}}

\begin{table}[]
\begin{minipage}{0.5\textwidth}
            \centering
\begin{tabular}{lcccc}
\hline
\multirow{2}{*}{Model} & \multicolumn{2}{c}{KG-Full} & \multicolumn{2}{c}{KG-50\%} \\ \cmidrule(lr){2-3} \cmidrule(lr){4-5}
                        & 1-hop         & 2-hop        & 1-hop         & 2-hop        \\ \hline
RoBERTa            & 90.2          & 70.8         & 61.5          & 39.3         \\ %\hline
RoB+G+M            & 91.4          & 72.6         & 62.5          & 40.8         \\ %\hline
JAKET               & \textbf{93.9}          & \textbf{73.2}         & \textbf{63.1}          & \textbf{41.9}         \\ \hline
\end{tabular}\caption{\label{tab:kgqa}Results on the MetaQA dataset over 1-hop and 2-hop questions under \textit{KG-Full} and \textit{KG-50\%} settings. RoB+G+M is the abbreviation for the baseline model RoBERTa+GNN+M.}
\end{minipage} \hfill
\begin{minipage}{0.45\textwidth}
\centering
\begin{tabular}{lccc}
\hline
   Model          & 100\% & 20\% & 5\%  \\ \hline
GNN          & 48.2  & -    & -    \\ %\hline
RoBERTa & 33.4  & -    & -    \\ %\hline
RoB+G+M   & 79.1  & 66.7 & 53.5 \\ %\hline
JAKET         & \textbf{81.6}  & \textbf{70.6} & \textbf{58.4} \\ \hline
\end{tabular}\caption{\label{tab:entclass}Results on the entity classification task over an unseen Wikidata knowledge graph. RoB+G+M is the abbreviation for the baseline model RoBERTa+GNN+M.}
\end{minipage}
\end{table}

\textbf{Entity Classification}. To further evaluate our model's capability to reason over unseen knowledge graphs, we design an entity classification task. Here, the model is given a portion of the Wikidata knowledge graph unseen during pre-training, denoted as $\mathcal{KG}'$. It needs to predict the category labels of these novel entities. The entity context embeddings are obtained in the same way as in pre-training. The relation context embeddings are generated by its surface text. The number of entities and relations in the $\mathcal{KG}'$ are 23,046 and 316 respectively. The number of triplets is 38,060. Among them, 16,529 entities have 1,291 distinct category labels.

We conduct experiments under a semi-supervised transductive setting by splitting the entities in $\mathcal{KG}'$ into train/dev/test splits of 20\%, 20\% and 60\%. To test the robustness of models to the size of training data, we evaluate models when using 20\% and 5\% of the original training set. 

In this task, RoBERTa takes the entity description text as input for label prediction while neglecting the structure information of KG. JAKET and RoBERTa+GNN+M make predictions based on the entity representation output from the knowledge module. We also include GNN as a baseline, which uses the same GAT-based structure as our knowledge module, but with randomly initialized model parameters and context embedding memory. GNN then employs the final entity representations for entity category prediction.% For RoBERTa, we directly use the initial pre-trained RoBERTa to generate entity context embeddings for label prediction. For RoBERTa+GNN+M, we use the output of knowledge module for label prediction.

As shown in Table~\ref{tab:entclass}, our model achieves the best performance under all the settings. The performance of GNN or RoBERTa alone is significantly lower than JAKET and RoBERTa+GNN+M, which demonstrates the importance of integrating both context and knowledge information using our proposed framework. Also, the gap between JAKET and RoBERTa+GNN+M increases when there's less training data, showing that the joint pre-training can reduce the model's dependence on downstream training data.

% \begin{table}[]
% \centering
% \begin{tabular}{lccc}
% \hline
%   Model          & 100\% & 20\% & 5\%  \\ \hline
% GNN          & 48.2  & -    & -    \\ %\hline
% RoBERTa & 33.4  & -    & -    \\ %\hline
% RoBERTa+GNN+M  & 79.1  & 66.7 & 53.5 \\ %\hline
% Ours         & \textbf{81.6}  & \textbf{70.6} & \textbf{58.4} \\ \hline
% \end{tabular}
% \caption{\label{tab:entclass}Results on Entity Classification.}
% \end{table}

\section{Conclusion}
\label{con}

This paper presents a novel framework, JAKET, to jointly pre-train models for knowledge graph and language understanding. Under our framework, the knowledge module and language module both provide essential information for each other. After pre-training, JAKET can quickly adapt to unseen knowledge graphs in new domains. Moreover, we design the entity context embedding memory which speeds up the pre-training by 15x.
Experiments show that JAKET outperforms baseline methods in several knowledge-aware NLU tasks: few-shot relation classification, KGQA and entity classification. In the future, we plan to extend our framework to natural language generation tasks.

% \subsubsection*{Acknowledgments}
% Use unnumbered third level headings for the acknowledgments. All
% acknowledgments, including those to funding agencies, go at the end of the paper.

\bibliography{iclr2021_conference}
\bibliographystyle{iclr2021_conference}

\appendix
\section{Appendix}
\subsection{Implementation details}
The dimension of hidden states in the knowledge module is 768, the same as $\text{RoBERTa}_{\texttt{BASE}}$, and the number of attention heads is 8. During pre-training, the batch size and length of text sequences are 1024 and 512 respectively. The batch size of KG entities are 16,384. The number of training epochs is 8. JAKET is optimized by AdamW~\citep{adamw} using the following parameters: $\beta_1=0.9$, $\beta_2=0.999$, $\epsilon=\text{1e-8}$, and weight decay of $0.01$. The learning rate of the language module is warmed up over the first 3,000 steps to a peak value of 1e-5, and then linearly decayed. The learning rate of our knowledge module starts from 1e-4 and then linearly decayed. 

\end{document}